\documentclass[runningheads,letterpaper]{llncs}
\usepackage{amsmath}

\usepackage{amssymb}
\usepackage{graphicx}

\usepackage{url}
\urldef{\mailsa}\path|{alfred.hofmann, ursula.barth, ingrid.haas, frank.holzwarth,|
\urldef{\mailsb}\path|anna.kramer, leonie.kunz, christine.reiss, nicole.sator,|
\urldef{\mailsc}\path|erika.siebert-cole, peter.strasser, lncs}@springer.com|    
\newcommand{\keywords}[1]{\par\addvspace\baselineskip
\noindent\keywordname\enspace\ignorespaces#1}

\newcommand{\tabspacing}{\vspace*{0cm}}

\begin{document}

\mainmatter  % start of an individual contribution

\title{Modeling the Temporal Nature of Human Behavior for Demographics Prediction}

\titlerunning{Modeling the Temporal Nature of Human Behavior} % too long

%P\aa{}l Sunds\o{}y\IEEEauthorrefmark{3},
%Alex `Sandy' Pentland\IEEEauthorrefmark{1}, 
%Sune Lehmann\IEEEauthorrefmark{2} and
%Yves-Alexandre de Montjoye\IEEEauthorrefmark{1}\IEEEauthorrefmark{4}}
%\IEEEauthorblockA{\IEEEauthorrefmark{1}MIT Media Lab, Cambridge, Massachusetts, USA}
%\IEEEauthorblockA{\IEEEauthorrefmark{2}Technical University of Denmark, Kgs. Lyngby, Denmark}
%\IEEEauthorblockA{\IEEEauthorrefmark{3}Telenor Research, Fornebu, Norway}
%\IEEEauthorblockA{\IEEEauthorrefmark{4}Imperial College London Dept. of Computing and Data Science Institute, London, UK}}

%\author{Bjarke Felbo^{1,3} \and P\aa{}l Sunds\o{}y^{2} \and Alex `Sandy' Pentland^{1} \and Sune Lehmann^{3} \and Yves-Alexandre de Montjoye^{1,4}}
\author{Bjarke Felbo\inst{1,3} \and P\aa{}l Sunds\o{}y\inst{2} \and Alex `Sandy' Pentland\inst{1} \and Sune Lehmann\inst{3} \and Yves-Alexandre de Montjoye\inst{1,4}}
\authorrunning{Modeling the Temporal Nature of Human Behavior}

\institute{Massachusetts Institute of Technology, MIT Media Lab
\and Telenor Research
\and Technical University of Denmark, DTU Compute
\and Imperial College London, Dept. of Computing and Data Science Institute}

%Tiergartenstr. 17, 69121 Heidelberg, Germany\\
%\mailsc}


\toctitle{Lecture Notes in Computer Science}
\tocauthor{Authors' Instructions}
\maketitle

\begin{abstract}

Mobile phone metadata is increasingly used for humanitarian purposes in developing countries as traditional data is scarce. Basic demographic information is however often absent from mobile phone datasets, limiting the operational impact of the datasets. For these reasons, there has been a growing interest in predicting demographic information from mobile phone metadata. Previous work focused on creating increasingly advanced features to be modeled with standard machine learning algorithms. We here instead model the raw mobile phone metadata directly using deep learning, exploiting the temporal nature of the patterns in the data. From high-level assumptions we design a data representation and convolutional network architecture for modeling patterns within a week. We then examine three strategies for aggregating patterns across weeks and show that our method reaches state-of-the-art accuracy on both age and gender prediction using only the temporal modality in mobile metadata. We finally validate our method on low activity users and evaluate the modeling assumptions.

\end{abstract}

\section{Introduction}

%\the\textwidth
%\the\textheight

For the first time last year, there were more active mobile phones in the world than humans~\cite{prepaid}. Every time one of these phones is being used to text or call, it generates mobile phone metadata or CDR (Call Detail Records). Collected at large scale this metadata -- records of who calls or texts whom, for how long, and from where -- provide a unique lens into the behavior of humans and societies. For instance, mobile phone metadata have been used to plan disaster response and inform public health policy~\cite{bengtsson2011improved,wesolowski2015impact}. The potential of mobile phone metadata is particularly high in developing countries where basic statistics such as population density or mobility are often either missing or suffer from severe biases~\cite{bias}. Last year, an expert advisory group to the United Nations emphasized the importance of mobile phone data in measuring and ultimately achieving the Sustainable Development Goals~\cite{UN}.

The potential of mobile phone data in developing countries has, however, been hindered by the absence of demographic information, such as age or gender, associated with the data. This issue has caused a growing interest in predicting demographic information from mobile phone metadata. While previous work has focused on developing increasingly complicated features, we here propose a novel way of modeling mobile phone metadata using deep learning. From high-level assumptions regarding the nature of temporal patterns, we design a data representation and convolutional network (ConvNet) architecture that reach state-of-the-art accuracy inferring both age and gender using only the temporal modality.

\begin{figure}[tp]
  \centering
  \includegraphics[trim=1.6cm 19.4cm 1.75cm 2.08cm, clip, width=0.8\columnwidth]{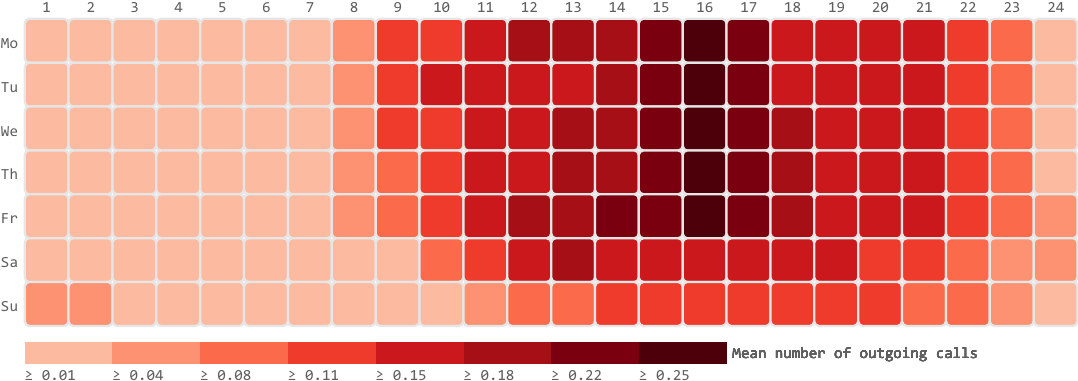}
  \caption{The mean number of outgoing calls averaged across the population. Differences between workdays and weekends are clearly visible as well as different times of the day.}
  \label{fig:punchcard}
\end{figure}

\section{Related work}
Previous work has relied heavily on hand-engineered features to predict demographics and other information from mobile phone metadata. Sarraute et al.~\cite{sarraute2014study} and Herrera-Yagüe et al~\cite{herrera2012prediction} both combined hand-engineered features with various machine learning algorithms to predict gender from mobile phone metadata while de Montjoye et al. used them to predict personality traits~\cite{de2013predicting}. Martinez et al. used an support vector machine (SVM) and random forest (RF) on similar features as well as a custom algorithm based on k-means to predict gender~\cite{frias2010gender}. Finally, Dong et al. used a double-dependent factor graph model to predict demographic information in a mobile phone social graph~\cite{dong2014inferring}. While promising, the graph-based approach requires demographic information about a large fraction of the population to be known a priori, making it impractical in most countries where training data is not available at scale and must be collected through surveys.

The current state of the art in predicting demographics from mobile phone data is a recent paper by Jahani et al.~\cite{Bandicoot} which relies on a large number of hand-engineered features (1440) provided by the open-source bandicoot toolbox~\cite{demontjoye2016bandicoot} and a carefully tuned SVM with a radial basis function kernel. The features used are divided into two categories (individual, spatial) and based on carefully engineered definitions such as how to group together calls and text messages into conversations or compute the churn rate of common locations.

\section{Data and Assumptions}

A mobile phone produces a record every time it sends or receives a text message or makes or receives a phone call. These records (called mobile phone metadata, or CDRs) are generated by the carrier's infrastructure and are highly standardized. CDRs contain the type of interaction (text/call), direction (in/out), timestamp (date and time), recipient ID, call duration (if call) and cell tower to which the phone was connected to. The dataset we work with, provided by an anonymous carrier, contains more than 250 million anonymized mobile phone records for 150.000 people in a Western European country covering a period of 14 weeks.

We state the following three assumptions about the nature of the temporal patterns in mobile phone metadata:

\medskip

\begin{enumerate}
    \setlength\itemsep{1em}
    \item \textbf{The day of the week and time of day of an observed \mbox{pattern} holds predictive power}\\ 
    
    Previous work showed that increasing the temporal granularity of the hand-engineered features in the bandicoot toolbox by differentiating between daytime and nighttime activity yields a substantial accuracy boost~\cite{Bandicoot}. For instance, the percentage of initiated calls at night during the weekend was one of the most useful features to predict gender.  Consequently, we assume that information on the specific time of the week that a pattern occurred contains useful information to predict demographic attributes.

    \item \textbf{Temporal patterns are similar across days of the week}\\
    
    While the time of day matters (e.g. night vs. day), we furthermore assume that such temporal patterns have similarities across days of the week which could help predict demographic attributes. For instance, one could imagine that a relevant temporal pattern on Friday night may help model a similar pattern on Saturday night. 

    \item \textbf{Local temporal patterns can be combined into predictive global features}\\
    
    The current state-of-the-art approach relies on complex hand-engineered (and non-linear) features such as the response rate within conversations, churn between antennas, and entropy of contacts~\cite{Bandicoot}. We assume that the convolutional network (ConvNet) will be able to combine local temporal patterns on the scale of hours to find global features (i.e. on the scale of days/weeks), thereby removing the need for such high-level hand-engineered features. ConvNets have similarly been used in previous work to learn a hierarchy of features directly from raw visual data~\cite{lecun2015deep}.

\end{enumerate}

\section{Representation, Architecture\\and Aggregation}

\subsection{Week-matrix representation}

Assumption 1 and Assumption 2 from Sect.~3 are used to derive our data representation for a week of mobile phone metadata. We represent the data as eight matrices summarizing mobile phone usage on a given week with hours of the day on the x-axis and the weekdays on the y-axis (see Figure~\ref{fig:punchcard}). These eight matrices are the number of unique contacts, calls, texts and the total duration of calls for incoming and outgoing interactions respectively. Every cell in the matrices  represents the amount of activity for a given variable of interest in that hour interval (e.g. between 2 and 3 pm). In this way, we effectively bin any number of interactions during the week. These eight matrices are combined into a 3-dimensional matrix with a separate `channel' for each of the 8 variables of interest. This 3-dimensional matrix is named a `week-matrix'.

The week-matrix representation is a logical result of our Assumption 1 and Assumption 2. Our first assumption focuses on the importance of high temporal granularity, which is why our data representation summarizes mobile phone usage for each hour, thereby splitting local patterns into separate bins such that they may be captured by a suitable classification algorithm. Our second assumption focuses on the similarity of temporal patterns across weekdays, making it logical to design the week-matrix to have the weekdays on the y-axis such that similar patterns are located in neighboring cells in the matrix (see Figure~\ref{fig:punchcard} for clear temporal patterns in mobile phone usage across weekdays). We shift the time in the matrices by 4 hours such that it is easier to capture mobile phone usage occurring across midnight (Figure~\ref{fig:punchcard} shows that there is especially a lot of activity occurring the night between Saturday and Sunday). Each row in the matrix thus contains data from 4am-4am instead of from midnight to midnight. This shift also moves the low-activity (and potentially less informative) areas to the borders of the matrix.

\subsection{ConvNet architecture}

We use our assumptions (see Sect.~3) to develop the ConvNet architecture used to model a single week of mobile metadata. The choice of architecture is crucial to finding predictive patterns and has been equated to a choice of prior~\cite{bengio2013representation}.

Assumption 2 emphasizes the similarity of temporal patterns across weekdays. We therefore design an architecture consisting of five horizontal conv.~layers followed by a vertical conv.~filter and a dense layer (see Table~\ref{tab:architecture} and Figure~\ref{fig:cnn_viz}). The horizontal conv.~layers learn to capture patterns within a single day, reusing the same parameters across different times of day and across the different weekdays. For a 1D conv.~filter with filter size four (as illustrated in Figure~\ref{fig:cnn_viz}) the value of a single neuron at the position $k$ in the next layer is:

\begin{equation}
o_{k} = \sigma \left(b + \sum_{l=0}^{3}w_{l}i_{k+l} \right),
\label{eqn:conv}
\end{equation}

where $w_l$ is position $l$ in the weight matrix for that filter and $b$ is the bias~\cite{michael2015}. The input is defined as $i_{k}$ for position $k$ in the previous layer. $\sigma$ is a non-linear activation function, which in this case is the leaky ReLU~\cite{maas2013rectifier}. A single conv.~layer consists of multiple filters with the specified size, allowing the conv.~layer to capture many different patterns across the entire input using only a few parameters.

The intraday patterns captured by the horizontal conv.~layers are then combined using the vertical conv.~layer across the different weekdays to find global features. Lastly, the dense layer and the softmax layer combine these global features to predict the demographic attribute (see Figure~\ref{fig:cnn_viz}).

Assumption 1 and 3 emphasizes the importance of capturing information about local temporal patterns. Consequently, we design an architecture that does not use pooling layers, which would throw away information about the location of the patterns in the week-matrix. Similarly, we make sure of a small conv.~filter size for the first four conv.~filter to focus on capturing local patterns.

There are many different parameters that can be tuned when choosing the architecture and the optimization procedure for training the ConvNet. Bayesian optimization is used for tuning seven of these as proposed in ~\cite{snoek2012practical}, covering e.g. the learning rate, L2 regularization, and the number of filters in the horizontal conv.~layers. The vertical conv.~layer has a fixed number of 400 filters. The dense layer has 400 neurons, whereas the softmax layer has as many neurons as the number of classes (two for gender and three for age).

\subsection{Aggregation of patterns across weeks}

The ConvNet architecture described models only a single week of data at a time, whereas each user has multiple weeks of data that should all be utilized when predicting a demographic attribute. Based on our three assumptions (see Sect.~3) it makes sense to design the ConvNet architecture to model a single week at a time, making it possible to reuse the same convolutional filters across multiple weeks. There are several ways to aggregate the features captured by the ConvNet for individual weeks, making our method utilize the data for multiple weeks. We examine three different approaches: averaging the predictions, adding a long short-term memory (LSTM) module to the ConvNet and modeling the features captured by the ConvNet with an SVM.

The most basic approach for modeling multiple weeks of data is to pass each week-matrix through the ConvNet architecture and then average the probabilities from the softmax layer. In this way, an overall prediction can be found across all weeks of data for a given user. An issue with this averaging approach is that it limits the contribution of a given week to the final prediction. 

\begin{table}[htp]
\caption{Architecture for the convolutional network. The filter size describes the number of neurons in the previous layer that each neuron in the current conv.~layer is connected to. A filter with size Mx1 takes as input M neurons located side-by-side horizontally, whereas a 1xN filter uses N neurons located side-by-side vertically.}

\label{tab:architecture}
\begin{center}
\begin{tabular}{|cc|}
\hline
Layer Name & Conv.~Filter Size \\
\hline
$input$ & - \\
$conv_1$ & 4x1 \\
$conv_2$ & 4x1 \\
$conv_3$ & 4x1 \\
$conv_4$ & 4x1 \\
$conv_5$ & 12x1 \\
$conv_6$ & 1x7 \\
$dense_7$ & - \\
$softmax_8$ & - \\
\hline
\end{tabular}
\end{center}
\end{table}
\tabspacing

\begin{figure}[tp]
\centering
\includegraphics[trim=4.6cm 1cm 3cm 2.0cm, clip, width=0.5\columnwidth]{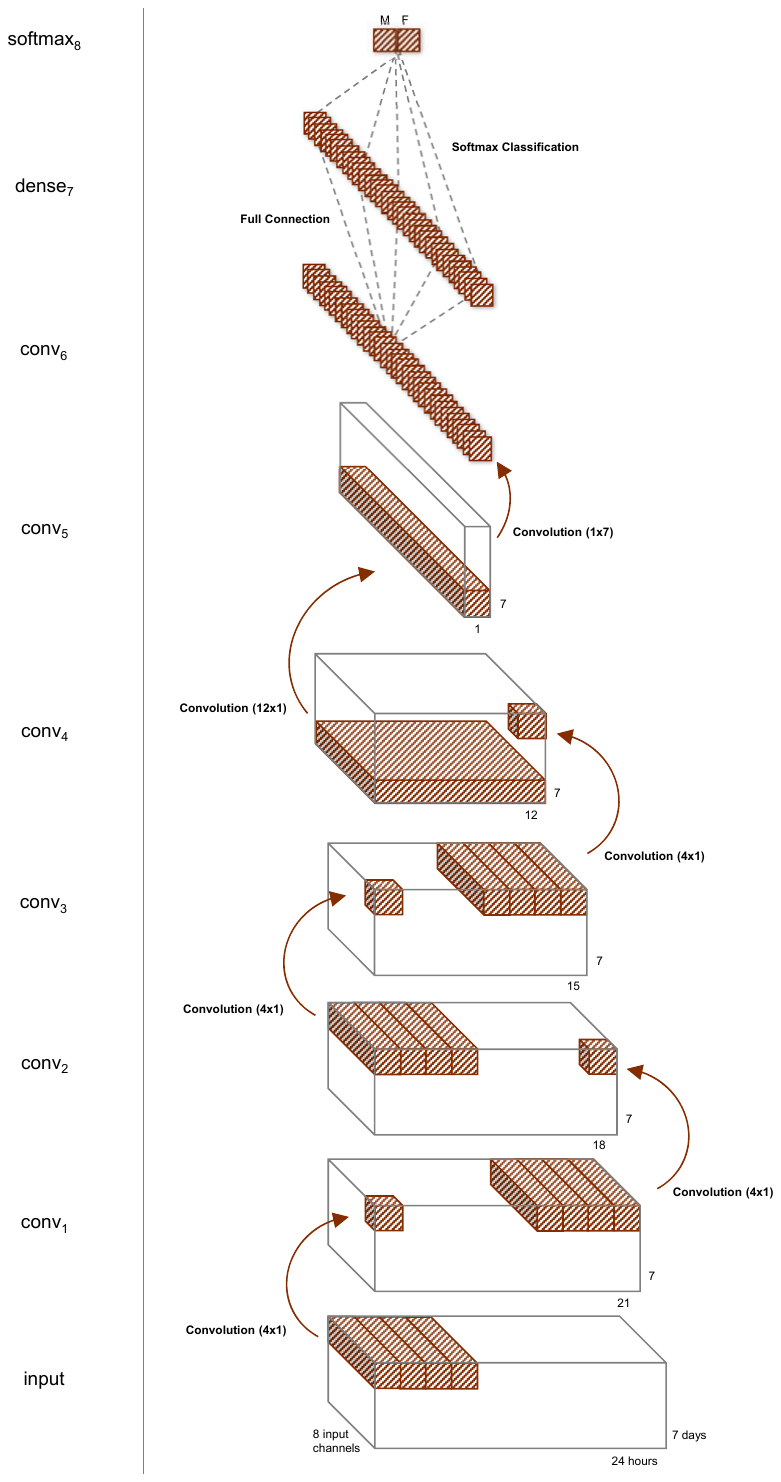}
\caption{Illustration of the convolutional network architecture. The depth of a conv.~layer equals the number of filters in that layer. Dimensions are not to scale.}
\label{fig:cnn_viz}
\end{figure}

Another way of modeling multiple weeks of data is by modifying the ConvNet architecture to include a long short-term memory (LSTM) module~\cite{hochreiter1997long}. The LSTM is a specialized variant of the recurrent neural network (RNN), which uses recurrent connections between the neurons to capture patterns in sequences of inputs. We design a ConvNet-LSTM such that it has the same architecture for finding patterns as our ConvNet architecture, but without the final softmax layer for classification (i.e. $conv_1$-$dense_7$ as seen in Figure~\ref{fig:cnn_viz}). This architecture is then connected to a 2-layer LSTM module with 128 hidden units in each layer. In this way, the week-matrices can be modeled with an end-to-end architecture that can utilize convolutional layers to find patterns within a week and recurrent layers to find patterns across weeks. It is trained using the default settings of the Adam optimization method~\cite{kingma2014adam}. L2 regularization of $10^{-4}$ and recurrent dropout~\cite{gal2015theoretically} of $0.5$ is used to avoid overfitting. The ConvNet-LSTM is implemented using Keras~\cite{chollet2015} and Theano~\cite{theano}.

Lastly, we use an SVM with a radial basis function kernel to design a 2-step model (ConvNet-SVM). The ConvNet is used to transform the raw data into learned high-level features for each week with the SVM then modeling patterns across weeks.  Using ConvNets to find good representations of raw data for modeling with SVMs has previously been done for generic visual recognition~\cite{donahue2013decaf}, but to our knowledge this is the first time it is done for combining patterns across individual observations in the dataset (i.e. weeks in this case). We extract the feature activations for $dense_7$ and $softmax_8$ (see Figure~\ref{fig:cnn_viz}. For each user we compute the mean and standard deviation for these extracted feature activations across the different weeks. A total of $800 + 2n_c$ features are extracted this way, where $n_c$ is the number of classes in the problem at hand (2 for gender, 3 for age). The number of features for the SVM is constant regardless of the number of weeks for a given user.

\section{Results}

In line with previous work and potential applications, we demonstrate the effectiveness of our method on gender and age prediction. We consider a binary gender variable (largest class: 56.3\%) and an age variable discretized by the data provider into three groups: $[18-39]$,$[40-49]$,$[50+]$, splitting the dataset almost equally (largest class: 35.7\%). Our dataset contains data of approximately 150.000 people. We split it into training (100.000 people), validation (10.000 people), and test set (40.000 people). We compare our results to a state-of-the-art approach, Bandicoot-SVM~\cite{Bandicoot}, using an SVM on the bandicoot features trained and tested on the same data as our method.

We report results using the three approaches for aggregating patterns across weeks described in Sect.~4. Table \ref{tab:exp_results} shows that our 2-step model (ConvNet-SVM), which extracts the high-level features found using the ConvNet and models them with an SVM yields the highest accuracy of the three approaches.

Our ConvNet-SVM method reaches state-of-the-art accuracy and slightly outperforms it on both age and gender prediction ($p < 10^{-5}$ with a one-tailed t-test). Our method reaches the state-of-the-art using only the temporal modality in mobile metadata, whereas the current state-of-the-art approach also exploits patterns related to mobility (see Sect.~7).

\begin{table}[h]
\caption{Accuracy of classifiers on the test set when predicting age and gender.}
\label{tab:exp_results}
\begin{center}
\begin{tabular}{lcc}
\hline
\itshape & Age & Gender \\
 \hline
 Random  & $35.7\%$ & $56.3\%$ \\
 %\hline
 Bandicoot-SVM & $61.6\%$ & $78.2\%$ \\
 %\hline
 ConvNet (averaging) & $60.7\%$ & $78.3\%$ \\
 %\hline
 ConvNet-LSTM & $61.3\%$ & $78.4\%$ \\
 %\hline
 ConvNet-SVM & $\mathbf{63.1\%}$ &  $\mathbf{79.7\%}$ \\
\hline
\end{tabular}
\end{center}
\end{table}
\tabspacing

%\subsection{Performance for low-activity users}

\begin{figure}[!htb]
    \centering
    \begin{minipage}[t]{.45\textwidth}
        \centering
        \includegraphics[trim=0cm 0cm 0cm 0cm, clip, width=1.0\columnwidth]{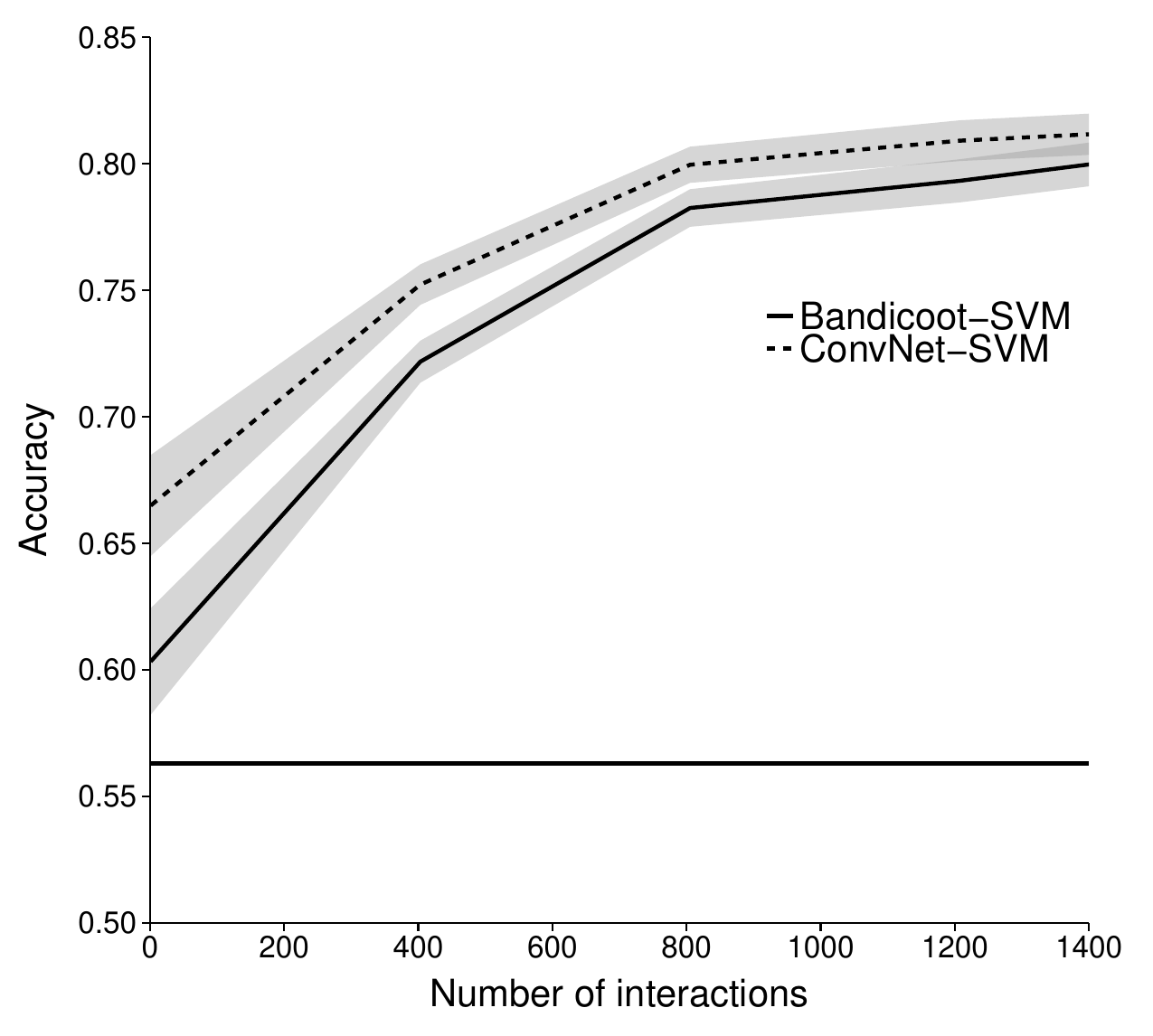}
\caption{Accuracy on gender prediction as a function of the number of interactions (across all 14 weeks) visualized using generalized additive model (GAM) smoothing. The x-axis is constrained to contain roughly 50\% of the users. The black solid line is the baseline accuracy when predicting everyone as part of the majority class.}
\label{fig:acc_few_int}
    \end{minipage}%
    \hspace{0.5cm}
    \begin{minipage}[t]{0.45\textwidth}
        \centering
        \includegraphics[trim=0cm 0cm 0cm 0cm, clip, width=1.0\columnwidth]{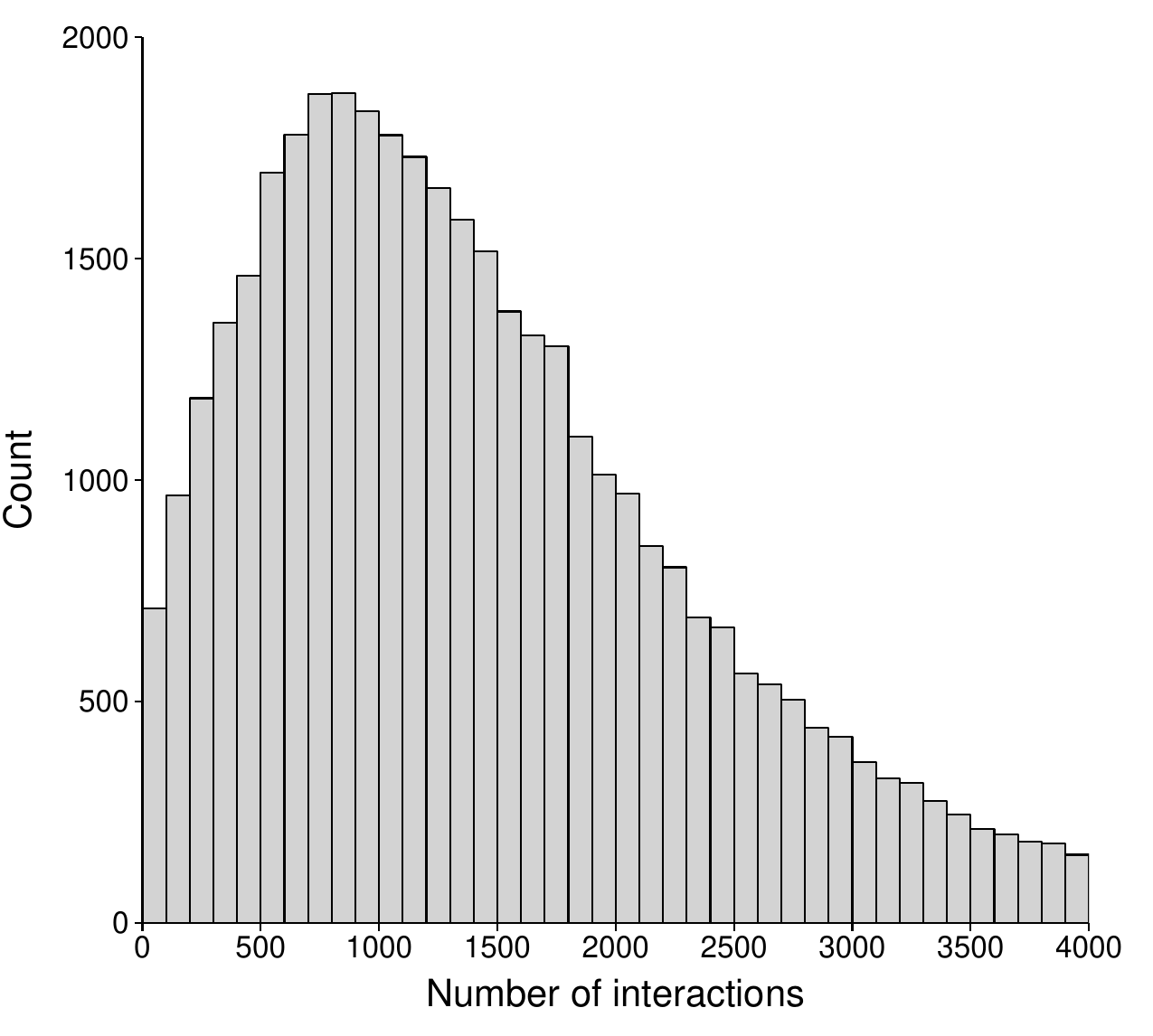}
\caption{Histogram of the distribution of the number of interactions. The top 5\% users in terms of number of interactions are not included.}
\label{fig:hist_interactions}
    \end{minipage}
\end{figure}

\begin{table}[!htp]
\caption{Accuracy on the original and the temporally randomized week-matrices.}
\label{tab:permutations}
\begin{center}
\begin{tabular}{lcc}
\hline
& Age & Gender\\
\hline
Original & 60.7 \% & 78.3\%\\
Permuted & 54.0 \% & 70.4\%\\  
\hline
Change & -11.0\% & -10.1\%\\
\hline
\end{tabular}
\end{center}
\end{table}
\tabspacing

% Future work: Find accuracy of age on low-activity users

Mobile phone usage in developing countries is still fairly low~\cite{prepaid} making it important for our method to perform well on low-activity users (see  Figure~\ref{fig:hist_interactions} for the distribution of interactions per user). To test the performance of our method, we train and evaluate it on low-activity users (users with fewer interactions than the median) and show that our model reaches state-of-the-art and even slightly outperforms it ($p < 0.01$ with a one-tailed t-test) with an accuracy of $76.9$\% vs. $75.7$\% for the Bandicoot-SVM. Figure~\ref{fig:acc_few_int} shows the accuracy of our method and the Bandicoot-SVM as a function of the number of interactions (calls + texts) when trained on all users showing that we perform particularly well on users with few interactions.

\section{Evaluating Assumptions}

Designing a ConvNet architecture for a particular modeling task involves many choices regarding filter sizes, layer types, etc. We derived many of our choices from the three assumptions stated in Sect.~3. In this section we evaluate these assumptions to qualify our choices.

\subsubsection{Evaluating Assumption 1:}

% Future work: Add temporal randomization by day and by 8 hour intervals.

The first assumption states that the weekday and time of day of an observed pattern holds predictive power. One way we can evaluate this assumption is by comparing the performance of a ConvNet on the original data with the performance of a ConvNet using the same hyperparameters and architecture but using data that has been temporally randomized. We temporally randomize the dataset by assigning values to cells at random in the week-matrix, thereby destroying potential temporal patterns in the week-matrices while keeping the rest of the information intact (total activity, etc). To quantify the impact of the temporal randomization independently of the SVM, we evaluate the performance when averaging predictions across weeks. Table \ref{tab:permutations} shows temporally randomizing the week-matrices decreases accuracy by 11\% when predicting age and by 10.1\% when predicting gender.

\begin{figure}[!hpt]
\centering
    \includegraphics[trim=0cm 1cm 0cm 0cm,clip, width=0.75\columnwidth]{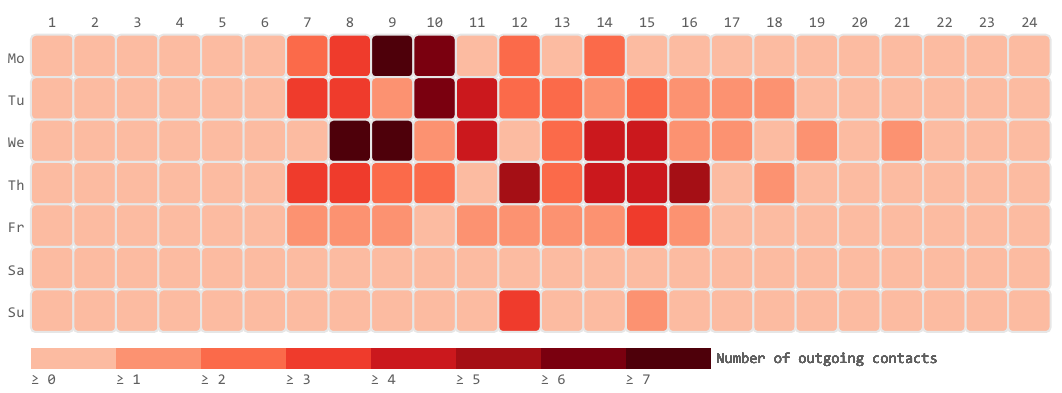}
    \vspace{0.2cm}
	  \includegraphics[trim=0cm 0cm 0cm 0.5cm, clip, width=0.75\columnwidth]{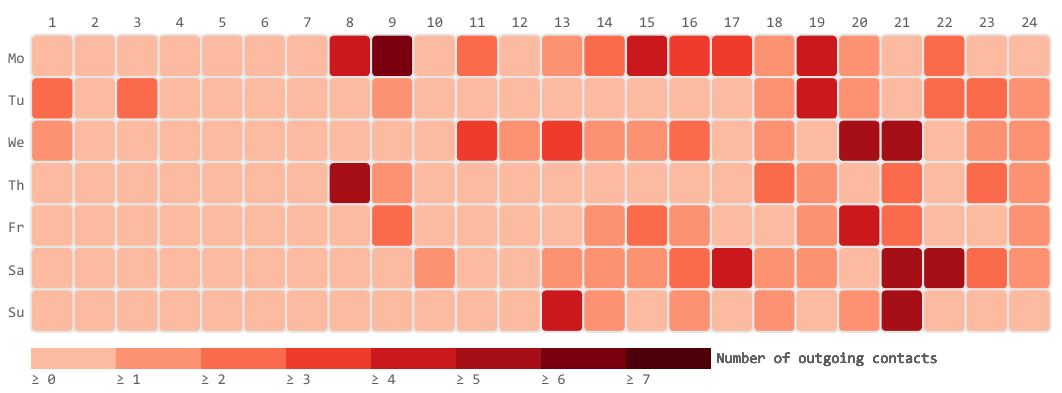}  
  \vspace{-0.3cm}
  \caption{Visualization of a single channel, the number of unique outgoing contacts, in the week-matrix most predictive of the male gender (top) and of female gender (bottom). The week-matrix most predictive of male gender has a higher number of outgoing contacts during the hours from 7am to 4pm on workdays while the  ``female'' week-matrix's outgoing contacts are spread across the day.}
  \label{fig:top_predictions}
\end{figure}

The importance of the time and day of the interactions is indicated by examining the week-matrices which our model is most confident belong to a man or a woman. Figure~\ref{fig:top_predictions} shows that the top ``men'' week-matrix has a higher number of outgoing contacts during the hours from 7 am to 4 pm on workdays while the top ``female'' week-matrix's outgoing contacts are spread across the day.

\subsubsection{Evaluating Assumption 2:}

The second assumption states that temporal patterns are similar across weekdays. To evaluate our assumption, we examine the performance of ConvNet architectures on a 1-dimensional representation of the data. While this 1D representation contains the same information as the week-matrix, the hours of the weekdays are arranged next to each other horizontally instead of vertically ($168 \times 1$ instead of $24 \times 7$, see Figure~\ref{fig:punchcard}) therefore preventing the ConvNet to exploit similarity in patterns across days of the week. We test multiple ConvNet architectures (examples in Table \ref{tab:1d_examples}) that have the same number of conv.~layers as our ConvNet architecture and a comparable number of parameters and show that all of these architectures yield a lower accuracy than our ConvNet and the current state-of-the-art approach.

\begin{table}[!hpt]
\caption{Examples of 1-dimensional ConvNet architectures that we have tested. These contain convolutional, dense, max-pool and softmax layers as denoted by the prefix. The filter size is shown in the suffix. The mark $(s)$ means that the conv.~layer has a stride of 2. Padding is used such that only pooling and stride reduces the dimensions.}
\label{tab:1d_examples}
\begin{center}
\begin{tabular}{|c|c|}
\hline
\textbf{ConvNet 1}  & \textbf{ConvNet 2} \\
\hline
\multicolumn{2}{|c|}{$input$} \\
\hline
$conv5$ & $conv13$ \\
$conv5$ & $conv13$ \\
$pool2$ & $conv13 (s)$ \\
$conv5$ & $conv13$ \\
$conv5$ & $conv13$ \\
$pool2$ & $conv13 (s)$ \\
$conv5$ & \\
$conv5$ & \\
\hline
\multicolumn{2}{|c|}{$dense$} \\
\hline
\multicolumn{2}{|c|}{$softmax$} \\
\hline
\end{tabular}
\end{center}
\end{table}
\tabspacing

\begin{table}[!hpt]
\caption{Top 5 bandicoot features captured by the neurons.}
\label{fig:bandi_histogram}

\centering
\begin{tabular}[h]{lc}
\textbf{Features} & $\lvert r \rvert$ \\
\hline
Interevent time (call) & 0.786 \\
Number of contacts (text) & 0.782 \\
Interevent time (text) & 0.769 \\
Entropy of contacts (call) & 0.764 \\
Number of interactions (text) & 0.761 \\
\hline
\end{tabular}
\end{table} 
\tabspacing

\subsubsection{Evaluating Assumption 3:}

The third assumption states that local temporal patterns captured by  convolutional filters (see Eqn.~\ref{eqn:conv}) can be combined into predictive global features, thereby eliminating the need for hand-engineered features. To evaluate this assumption, we examine the global features learned with our deep learning method by comparing the patterns captured by the neurons of our ConvNet\footnote{For this comparison we use the mean activation of neurons in the $FC_7$ layer.} with the bandicoot features. We only consider the individual bandicoot features as our ConvNet does not capture location and movement information used for the mobility features.

Table~\ref{fig:bandi_histogram} shows that the ConvNet captures information very similar to the one encoded in high-level hand-engineered features such as interevent time and entropy of contacts, suggesting that our deep learning model combines local temporal patterns into global features.

\section{Discussion}

Our results (Table~\ref{tab:exp_results}) show that the ConvNet-SVM outperforms the ConvNet-LSTM despite the ConvNet-SVM not capturing the ordering of the week-matrices. While an in-depth study is outside the scope of this paper, these results suggest that there are no strong inter-week patterns that are crucial for predicting demographic attributes. 

The state-of-the-art approach found that two mobility features (percent interactions at home and entropy of antennas) were among the top 5 most predictive features for one of their two benchmark datasets~\cite{Bandicoot}. In contrast, our ConvNet-SVM method reached state-of-the-art accuracy despite not using mobility information at all. In future work, we would like to use deep learning methods for modeling the other modalities in mobile phone metadata as well, thereby likely increasing the prediction accuracy. Our weekmatrix representation have been added to bandicoot\footnote{Version $\ge$ 0.4 at \texttt{http://bandicoot.mit.edu} under \texttt{bc.special.punchcard}} and our trained ConvNets for Caffe~\cite{jia2014caffe} are available\footnote{\texttt{https://github.com/yvesalexandre/convnet-metadata/}}.

%\section*{Acknowledgment}

%We would like to thank Lars Kai Hansen at the Technical University of Denmark (DTU) for useful discussions. This work was supported in part by the Army Research Laboratory under Cooperative Agreement Number W911NF-09-2-0053. Sune Lehmann acknowledges support from the Villum Foundation and the Danish Council for Independent Research. Yves-Alexandre de Montjoye was partially supported by a grant from the Media Lab and Wallonie-Bruxelles International.

\bibliography{references}
\bibliographystyle{splncs03}

\end{document}